\documentclass{article}

\usepackage{PRIMEarxiv}
\usepackage[utf8]{inputenc} % allow utf-8 input
\usepackage[T1]{fontenc}    % use 8-bit T1 fonts
\usepackage{hyperref}       % hyperlinks
\usepackage{url}            % simple URL typesetting
\usepackage{booktabs}       % professional-quality tables
\usepackage{amsfonts}       % blackboard math symbols
\usepackage{nicefrac}       % compact symbols for 1/2, etc.
\usepackage{microtype}      % microtypography
\usepackage{lipsum}
\usepackage{fancyhdr}       % header
\usepackage{graphicx}       % graphics
\graphicspath{{media/}}     % organize your images and other figures under media/ folder
\usepackage{filecontents}
\usepackage{amsmath,amssymb,amsfonts}
\usepackage{algorithmic}
\usepackage{textcomp}
\usepackage{xcolor}
\usepackage[most]{tcolorbox}
\usepackage{framed}
\definecolor{shadecolor}{RGB}{0,200,230}
\usepackage{graphicx}
\usepackage{adjustbox}
\title{Geotokens and Geotransformers
}

\author{
  Eren Unlu \\
  Paris, France\\
  \texttt{eren.unlu@matyz.org} \\
   \\
}

\begin{document}
\maketitle

\begin{abstract}
In transformer architectures, position encoding primarily provides a sense of sequence for input tokens. While the original transformer paper's method has shown satisfactory results in general language processing tasks, there have been new proposals, such as Rotary Position Embedding (RoPE), for further improvement. This paper presents geotokens, input components for transformers, each linked to a specific geological location. Unlike typical language sequences, for these tokens, the order is not as vital as the geographical coordinates themselves. To represent the relative position in this context and to keep a balance between the real world distance and the distance in the embedding space, we design a position encoding approach drawing from the RoPE structure but tailored for spherical coordinates. 
\end{abstract}

% keywords can be removed

\section{Introduction}
The transformer model, introduced by \cite{vaswani2017attention}, has established itself as an efficient framework, becoming central to many cutting-edge natural language generation tasks and broader generative AI \cite{mauricio2023comparing}, \cite{wei2022emergent}. Unlike its forerunners, such as Recurrent Neural Networks (RNNs), transformers don't inherently encode sequence order due to their parallel processing nature with self-attention \cite{adrian2020visualizing}, \cite{schmidt2019recurrent}. To address this, Vaswani and colleagues incorporated position encodings, which has since been fundamental for many AI models.
Though the original transformer's position encoding has been effective in understanding sequence order \cite{vaswani2017attention}\cite{yun2019transformers}, researchers have continuously strived to refine it \cite{shaw2018self}\cite{ke2020rethinking}. Proposals include encoding relative token positions and enhancing their interaction with matrices, though they often challenge compatibility with the linear self-attention structure as highlighted by \cite{su2021roformer}.

The recent innovation, Rotary Position Embedding (RoPE), offers a fresh perspective on incorporating positional data into transformers \cite{su2021roformer}. Unlike conventional techniques, RoPE uses a rotation matrix to simultaneously capture relative position information within the self-attention process. This paper expands on the RoPE concept, adjusting it for a three-dimensional geographical space, vital for geographic data representation.

We delve into 'geotokens' and a cartographical transformer framework called 'geotransformers'. Efficiently encoding geographic entities opens doors for the next AI generation, focusing on the accurate representation of physical locations while preserving relative distances and hierarchies. Thus, we propose a transformer model where each input is a 'geotoken' symbolizing a geographic element. As discussed, geotokens emphasize spatial relations over sequences. The primary consideration of a geotoken isn't its sequence placement, like traditional language tokens, but its geographic coordinates and relation to other geotokens. To represent this geographic positioning, we suggest a three-dimensional adaptation of the RoPE model for spherical coordinates. The efficiency of the proposed model is demonstrated under a proper experimental setting.

\section{Notion of Geotokens and Geotransformers}
Encoding geographical entities holds immense promise, particularly as we move into an age where data is not only textual but also spatial. The insights garnered from such data can revolutionize various fields, from urban development and environmental tracking to navigation and tourism. While some suggest integrating spatial encoding into natural language, as seen in \cite{unlu2023chatmap}, there's a clear need for a more effective and sturdy approach to encode geographical coordinates. As our world becomes increasingly interconnected and reliant on precise geospatial data, the mechanisms by which we interpret and represent this data will be paramount in determining our ability to innovate and progress in myriad disciplines.

This is where transformer architectures come into play. Originally designed for natural language processing tasks, transformers have demonstrated remarkable adaptability and efficiency in handling sequences. Their self-attention mechanism allows them to weigh the importance of different elements in a sequence relative to each other, making them particularly suitable for spatial data where relationships and relative positions between entities can be crucial. For instance, in a geographical setting, understanding the proximity or relative positioning of various landmarks can be more informative than merely knowing their individual locations. Moreover, the parallel processing capabilities of transformers enable them to handle vast amounts of data simultaneously. This is particularly beneficial for spatial datasets that often come with a multitude of variables, such as elevation, temperature, or land usage type. Transformer architectures can seamlessly integrate these variables, providing a holistic understanding of the geographical space. Furthermore, the flexibility of transformers allows for the integration of various data modalities. In the context of geography, this means that textual descriptions, satellite imagery, and raw coordinate data can all be processed together, creating a comprehensive representation of the geographical entity.

In this study, we present a novel approach by devising a transformer-based architecture tailored not for conventional textual tokens but for "geotokens." A geotoken is a sophisticated data entity that not only carries the semantic significance but also encodes the spatial attributes of a geographical unit. These geotokens can be representative of diverse geographical elements, from distinct landmarks like renowned research facilities or observatories to expansive zones such as tectonic plate boundaries or metropolitan regions. To streamline our initial approach, we focus primarily on discrete locations characterized by their specific latitude and longitude coordinates. It's pertinent to note that each of these geotokens is backed by a pre-embedded vector that captures vital information about the geographical entity in question. This information could have been procured through various pre-trained neural architectures or mechanisms, be it a natural language processing model delving into its descriptive text or a Convolutional Neural Network (CNN) distilling its visual information.

In our discussion, a "geotoken" signifies a tokenized portrayal of any geographical element, which might encompass distinct landmarks, locales, or even vast territories. In contrast to traditional tokens that depict words or symbols within a text, geotokens encapsulate spatial characteristics, the inherent meaning, and the surrounding context of these geographical features. When we envision a transformer model designed to process geotokens, it becomes evident that it shouldn't merely be encoding their sequence but rather their specific geographical coordinates. This realization underscores the need for a revamped position encoding methodology tailored for this context. Recognizing this, researchers are now diving deeper into crafting specialized transformer models that can optimally handle geotokens. With the rise of spatial data and geospatial analytics, there's an imminent need to align machine learning models, especially transformers, with these requirements. As the digital landscape evolves, understanding and integrating such geospatial nuances into AI models will become increasingly paramount.

\begin{figure}[h]
\centering
\label{fig:fig_2}
\includegraphics[width=0.65\linewidth]{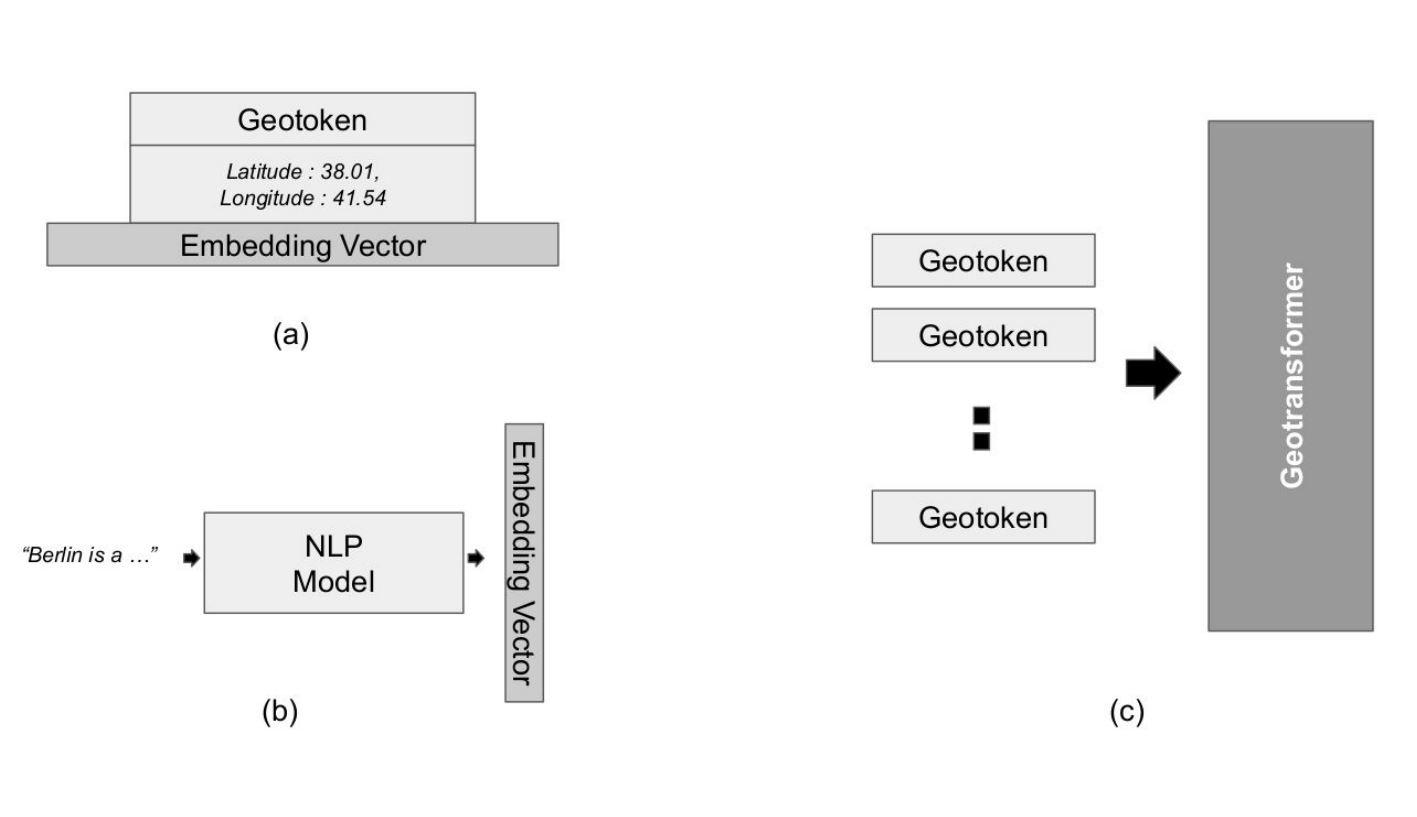}
\caption{(a) A basic geotoken is characterized by its location, given in terms of latitude and longitude. (b) The underlying features of this geotoken could be encoded using any pre-existing neural model, in this instance, a textual description processed through NLP. (c) The Geotransformer framework is designed to handle these geotokens.}
\end{figure}

\section{Original Position Encoding Mechanism}
\label{Original Position Encoding Mechanism}

As highlighted earlier, the method suggested for encoding sequential positions in the initial transformer architecture has demonstrated its effectiveness, even with its minor limitations. Adopting the same conventions as in \cite{su2021roformer}, let $\mathbb{S} = \{w_{i}\}^{N}_{i=1}$, for $N$ input tokens with input embedding vectors of $\mathbb{E} = \{x_{i}\}^{N}_{i=1}$. The projected query, key, value vectors in self-attention are as follows respectively :

\begin{equation} \label{eq1}
\begin{aligned} 
q_{m} = f_{q}(x_{m}, m)\\
k_{n} = f_{k}(x_{m}, n)\\
v_{n} = f_{v}(x_{m}, n)\\
\end{aligned} 
\end{equation}

$m$ and $n$ denoting the respective positions in vectors. 
\cite{vaswani2017attention} suggests introducing non-trainable additive matrices to the input embeddings, described as:

\begin{equation} \label{eq2}
\begin{aligned} 
p_{i,2t} = sin(k/10000^{2t,d})\\
p_{i,2t+1} = cos(k/10000^{2t,d})\\
\end{aligned} 
\end{equation}

where sine and cosine functions of absolute token position is encoded in even and odd numbered indexes of positional vector of same size with the $d$ dimensional input embeddings. The intuitive theory behind this is that through trigonometric identity functions the relative positional distances can be represented by linear algebraic multiplications of the encodings.

\section{Rotary Position Encoding (RoPE)}

\cite{he2020deberta} posited a perspective on how to model the relative positions of two tokens. Subsequently, \cite{su2021roformer} initially presented the position encoding as a function $g$, which operates on the inner products of query and key vectors : 

\begin{equation} \label{eq3}
\begin{aligned} 
 \langle f_{q}(x_{m},m), f_{k}(x_{n},n)\rangle = g(x_{m},x_{n},m-n)\\
\end{aligned} 
\end{equation}

$m-n$ representing the relative positions. Via this type of a basis formulation authors prove the relative position can be encoded as a rotation matrix as :

\begin{equation}\label{eq4}
\begin{aligned} 
f_{ \{q,k\} }(x_{m},m) = \mathbf{R}_{\Theta, m}^{d} \mathbf{W} x_{m}\\
\end{aligned} 
\end{equation}

$\mathbf{R}_{\Theta, m}^{d}$ being the proposed rotation matrix as in (5) where inner sections are taken from regular two dimensional matrix rotation. From the rotation matrix, it's evident that the embedding space is evenly divided, with the positional indices integrated within. This approach seems to draw inspiration from the original absolute position encoding, with $\theta$ serving a role akin to the angle function as in (6) :

\begin{equation}\label{eq5}
\resizebox{.8\hsize}{!}{$
\mathbf{R}_{\Theta, m}^{d} = 
\begin{bmatrix} 
    cos(m\theta_{1}) & -sin(m\theta_{1}) & 0  & 0 & \dots  & 0  & 0 \\
    sin(m\theta_{1}) & cos(m\theta_{1}) & 0  & 0 & \dots   & 0  & 0 \\
    0  & 0 & cos(m\theta_{2}) & -sin(m\theta_{2}) & \dots  & 0  & 0 \\
    0  & 0 & sin(m\theta_{2}) & cos(m\theta_{2}) & \dots  & 0  & 0 \\
    \vdots & \vdots & \vdots & \vdots & \ddots & \vdots & \vdots \\
     0  & 0 &  0  & 0 & \dots & cos(m\theta_{d/2}) & -sin(m\theta_{d/2}) \\
     0  & 0 &  0  & 0 & \dots & sin(m\theta_{d/2}) & cos(m\theta_{d/2})
    \end{bmatrix}
$} 
\end{equation}

\begin{equation} \label{eq6}
\begin{aligned} 
\theta_{i} = 10000^{-(2i-1)/d}
\end{aligned} 
\end{equation}

\section{Spherical Position Encoding}

As previously highlighted, in a framework where geotokens are position-encoded based on their global coordinates, there's a requirement for a mechanism that accommodates spherical space. To address this, we suggest adapting the RoPE technique to work with spherical coordinates. We can denote the longitude and latitude of any given position as $\theta$ and $\phi$, respectively. For ease of understanding and without compromising the broader concept (while overlooking minor errors), let's make the assumption that our globe is an ideal sphere with a consistent radius, $R$.

Euler angles in three dimensions can describe the rotation matrix within this stable coordinate framework. The rotation matrix's standard representation in this context is outlined below in (7), $\phi$, $\psi$, $\theta$ denoting rotation along $x$, $y$, $z$ axes respectively.

\begin{equation}\label{eq5}
\resizebox{.92\hsize}{!}{$
\mathbf{R}_{e} = 
\begin{bmatrix} 
    cos(\psi)cos(\theta) & -cos(\phi)sin(\theta)+sin(\phi)sin(\psi)cos(\theta) & sin(\phi)sin(\theta)+cos(\phi)sin(\psi)cos(\theta) \\
     cos(\psi)sin(\theta) & cos(\phi)cos(\theta)+sin(\phi)sin(\psi)sin(\theta) & -sin(\phi)cos(\theta)+cos(\phi)sin(\psi)sin(\theta) \\ 
     -sin(\psi) & sin(\phi)cos(\psi) & cos(\phi)cos(\psi) \\ 
    \end{bmatrix}
$} 
\end{equation}

Assuming longitude and latitude variation defined as a rotation on a sphere along $x$ and $z$ axes respectively with setting the angles $\phi$ and $\theta$, intuitively we need to keep the $y$-axis rotation constant by equating the angle $\psi = 0$. Therefore the rotation matrix in this case can be written as in (8).

\begin{equation}\label{eq5}
\resizebox{.5\hsize}{!}{$
\mathbf{R}_{e}^{'} = 
\begin{bmatrix} 
    cos(\theta) & -cos(\phi)sin(\theta) & sin(\phi)sin(\theta) \\
     sin(\theta) & cos(\phi)cos(\theta) & -sin(\phi)cos(\theta) \\ 
     0 & sin(\phi)  & cos(\phi)  \\ 
    \end{bmatrix}
$} 
\end{equation}

Basing on this particular rotation matrix (8), taking inspiration from the RoPE architecture \cite{su2021roformer}, we propose to encode a rotational position encoding matrix as follows, denoted by $\mathbf{R}_{\phi, \theta}^{d}$, which adhere to previously mentioned spherical geometry of two varying angles. We assume an embedding dimension divisible by three without loss of generality. The $\phi$ and $\theta$ in the proposed rotation matrix correspond to longitude and latitude in radial values of a given geotoken. 

\begin{equation}\label{eq5}
\resizebox{.92\hsize}{!}{$
\mathbf{R}_{\phi, \theta}^{d} = 
\begin{bmatrix}

    cos(\theta) & -cos(\phi)sin(\theta)  & sin(\phi)sin(\theta) & 0  & 0 & 0 & \dots & 0 & 0  & 0 \\
    sin(\theta) & -cos(\phi)cos(\theta)  & -sin(\phi)cos(\theta) & 0  & 0 & 0 & \dots & 0 & 0  & 0 \\
    0 & sin(\phi)  & cos(\phi) & 0  & 0 & 0 & \dots & 0 & 0  & 0 \\

    0 & 0  & 0 & cos(\theta) & -cos(\phi)sin(\theta)  & sin(\phi)sin(\theta) & \dots & 0 & 0  & 0 \\
    0 & 0  & 0 & sin(\theta) & -cos(\phi)cos(\theta)  & -sin(\phi)cos(\theta) & \dots & 0 & 0  & 0 \\
    0 & 0  & 0 & 0 &  sin(\phi)  & cos(\phi) & \dots & 0 & 0  & 0 \\
    
    \vdots & \vdots & \vdots & \vdots  & \vdots   & \vdots  & \ddots & \vdots & \vdots & \vdots \\
    
    0 & 0  & 0 & 0 & 0 & 0  & \dots & cos(\theta) & -cos(\phi)sin(\theta)  & sin(\phi)sin(\theta) \\
    0 & 0  & 0 & 0 & 0 & 0  & \dots & sin(\theta) & -cos(\phi)cos(\theta)  & -sin(\phi)cos(\theta) \\
    0 & 0  & 0 & 0 & 0 & 0  & \dots & 0 & sin(\phi)  & cos(\phi)   \\

    \end{bmatrix}
$} 
\end{equation}

Note that we do not need to calculate an auxiallary angle value as in original absolute position encoding or RoPE as geographical coordinates are inherently angular. For clarity, we've chosen an embedding dimension that's a multiple of three based on inherent requirements. However, this decision might not be ideal since many embedding mechanisms from different modalities may not conform to this restriction. While there are potential workarounds, like adding padding indices, they are beyond the purview of this paper. Additionally, other potential challenges like appropriate scaling are also not covered here, particularly when the architecture is trained with a limited set of geolocations instead of global data.

\section{Experimental Results}

In order to test the efficacy of the proposed encoding mechanism we have set an exemplary proof-of-concept framework. As such a geo-encoding based transformer architecture is being proposed for the first time, benchmarking against a particular transformer architecture or a position encoding method applicable for language processing context is not possible. Therefore, a new experiment setting  is needed which measures the proposed position encoding method fairly and accurately.

The experimental framework consists of predicting the distance from a given latitude, longitude in text, for a latitudinal and longitudinal displacement. Hence, the setting resembles to the textual sequence forecasting of regular language models, but with a limited vocabulary and token formulation. The model expects text in input in the form such as $'46.4157,21.0756+-0.0424,0.0132'$, which represents a given latitude and longitude and the coordinate displacements. The output is the distance in meters such as $'36224.102'$. The vocabulary consists of single characters with full set of $'0123456789.,+'$, in addition to regular special tokens like 'End-of-Sentence'.

For this context, the geotransformer is benchmarked against an other instance with same hyperparameters where proposed spherical position encoding is done with randomly fabricated latitudes and longitudes. The true enconding mechanism shall improve the training under these circumstances compared to a random encoding as the locations are properly embedded in the latent space with respecting relative spherical distances and much better alignment between textual character processing and coordinates is attained.

\begin{figure}[h]
\centering
\label{fig:fig_2}
\includegraphics[width=0.65\linewidth]{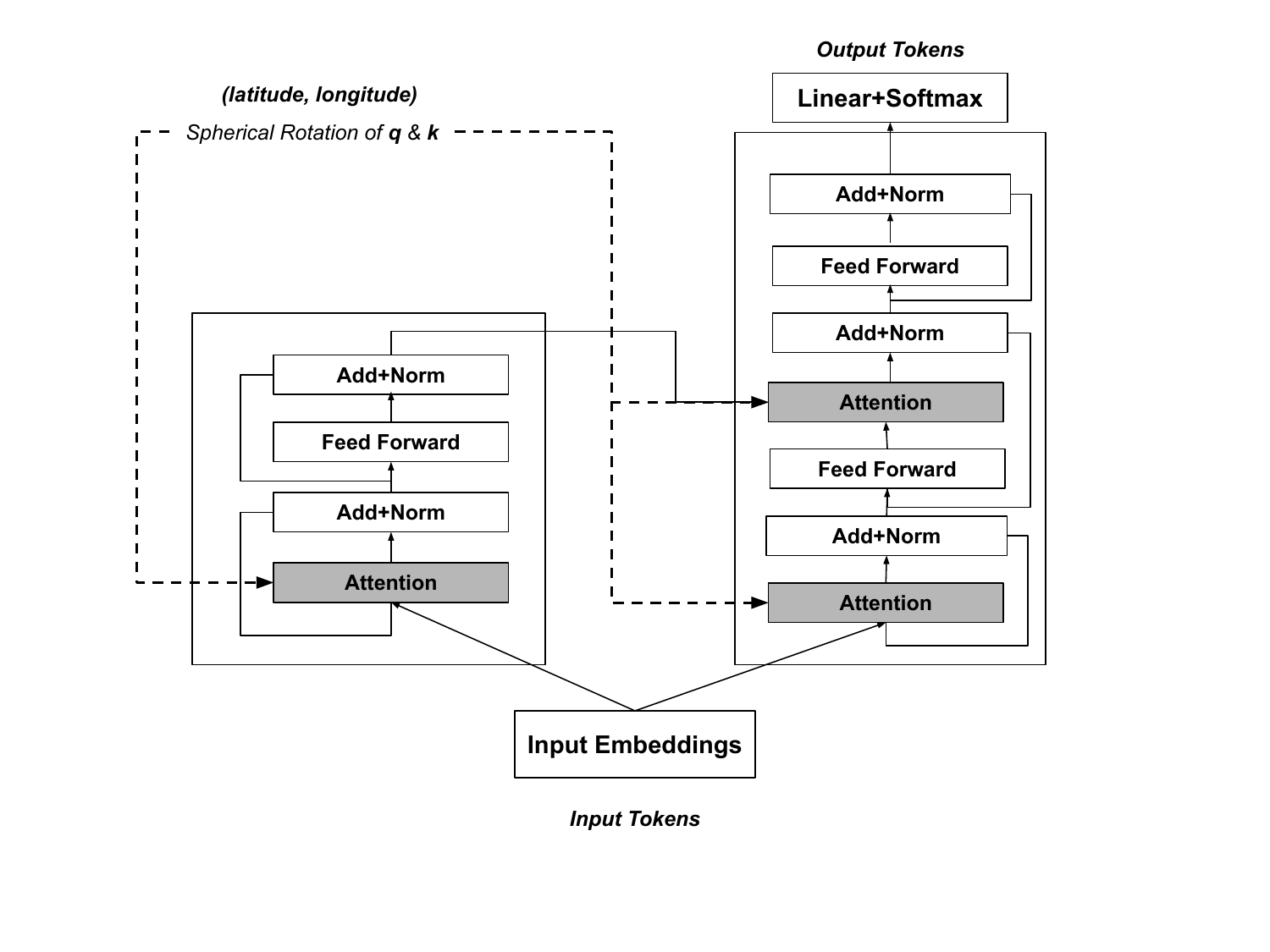}
\caption{The illustration of the used geotransformer architecture in the experiment.}
\end{figure}

A regular encoding-decoding transformer architecture \cite{vaswani2017attention} is used along with the proposed spherical rotation matrix for query and key vectors. It consists of a single encoder-decoder block with a single attention head. Encoding and Decoding blocks are fed with same input without masks as an unbiased reference to predict the outputs. Embedding dimension of tokens is 27, a multiple of 3 which conforms to the spherical rotation matrix. However, note that as previously mentioned, by proper padding or interleaving any embedding dimension can be used. Maximum sequence length is set to 100. Spherical rotation is applied on all attention layers including self-attention and encoder-decoder attention. Note that, no position encoding is applied for the tokens.

The dataset pairs consist of 512 random latitudes and longitudes on earth with uniformly random latitudinal, longitudinal displacements lower than 10 degrees. Regular Adam optimization is used with a learning rate of $10{-4}$ with beta values of 0.9 and 0.98. Both settings are trained for 25 epochs with a batch size of 64.

\begin{figure}[h]
\centering
\label{fig:fig_3}
\includegraphics[width=0.65\linewidth]{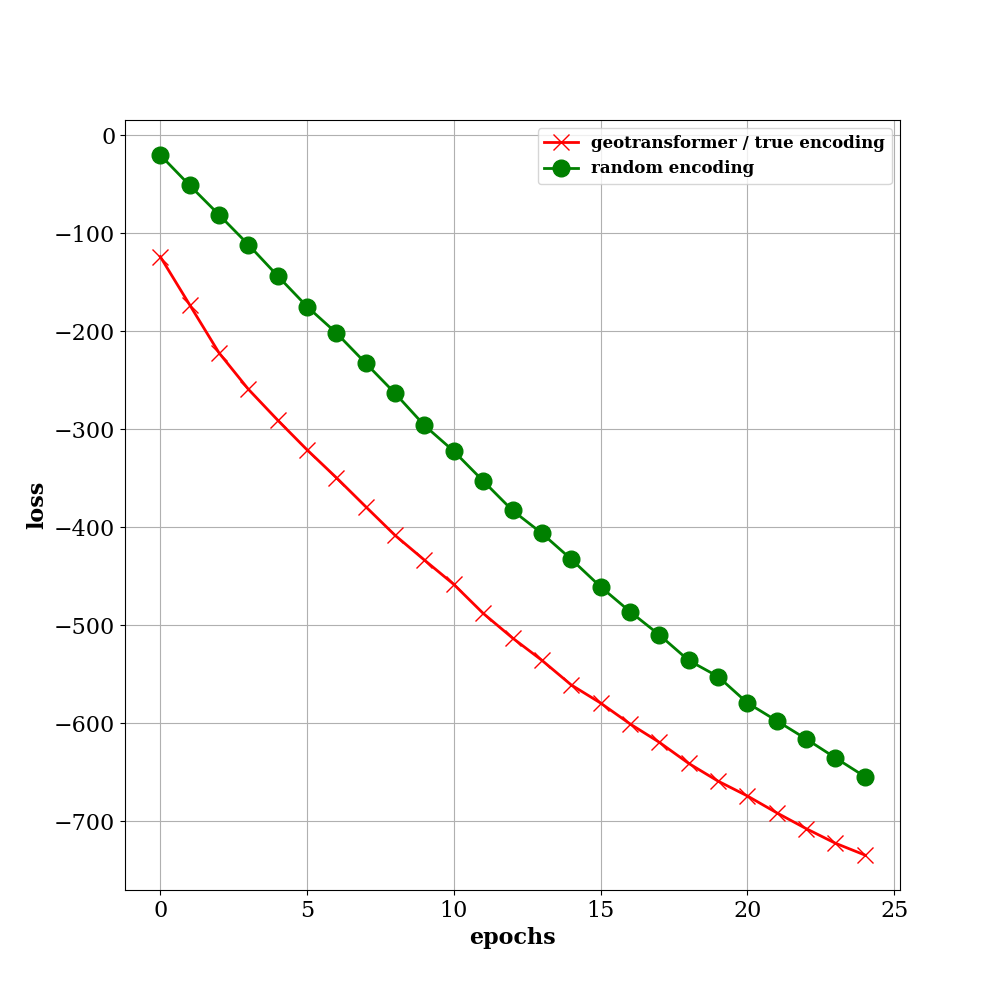}
\caption{Training loss of properly geo-encoded locations versus random latitudes and longitudes for the same geotransformer architecture for the proposed experimental setting. Spherical position encoding improves the training process significantly, which constitutes an early proof-of-concept.}
\end{figure}

Regular cross-entropy of token prediction probabilities is used as a training loss. Fig. 3 shows the total training loss per epoch. Proposed spherical position embedding significantly decreases the training losses compared to a case where random locations are geo-encoded, which constitutes an important early proof-of-concept. By properly rotating the query and key vectors respecting the spherical displacements, the model is able to effectively embed the notions of geolocations and proximity in the latent space. These results indicate that these kind of geospatially aware geometric rotations can be used in further advanced architectures. 

\section{Conclusion}

In our research, we've unveiled a groundbreaking adaptation of the transformer architecture that incorporates "geotokens", which serve as representations of distinct geographical entities. This innovative approach enables us to represent specific geographical locations through their associated feature vectors. The efficiency of this approach lies in its versatility, allowing these feature vectors to be potentially encoded using various pre-trained neural models across different modalities. However, this fusion isn't merely an augmentation in semantics. It also propels us into new terrains of challenge, notably the need to encode actual geographical coordinates instead of the conventional sequential positioning we've grown accustomed to. Acknowledging the inadequacies of traditional position embeddings when dealing with this spatial dimension, we adopted and further refined the Rotary Position Embedding (RoPE) technique. The efficiency of the proposed encoding mechanism to augment a transformer based model's cognition of geospatial concepts such as relative proximity is demonstrated with an experiment. This refined technique has been tailored specifically to cater to spherical coordinates, ensuring that the transformer architecture is finely attuned and harmonized with geographical data intricacies. We believe this advancement might pave the way for more accurate and contextually aware geographical data processing with transformer based architectures.

%%%%%%%%%%%%%%%%%%%%%%%%%%%%%%%%%%%%%%%%%%%%%%%%%%%%%%%%%%%%%%%%%%%%%%
%%%%%%%%%%%%%%%%%%%%%%%%%%%%%%%%%%%%%%%%%%%%%%%%%%%%%%%%%%%%%%%%%%%%%%

\end{document}